\documentclass[10pt,twocolumn,letterpaper]{article}

\usepackage{cvpr}              

\definecolor{cvprblue}{rgb}{0.21,0.49,0.74}
\usepackage[pagebackref,breaklinks,colorlinks,allcolors=cvprblue]{hyperref}
\usepackage{booktabs}
\usepackage{times}

\def\paperID{11620} 
\def\confName{CVPR}
\def\confYear{2025}

\newcommand{\yr}[1]{\textcolor{black}{#1}}

\newcommand{\revised}[1]{\textcolor{black}{#1}}
\newcommand{\yrn}[1]{\textcolor{black}{#1}}

\title{3D Gaussian Head Avatars with Expressive Dynamic Appearances by \\ Compact Tensorial Representations}

\begin{document}
\author{
Yating Wang$^1$ \and
Xuan Wang$^2$ \and
Ran Yi$^{1}$ \and 
Yanbo Fan$^2$ \and
Jichen Hu$^1$ \and
Jingcheng Zhu$^1$ \and 
Lizhuang Ma$^{1*}$ 
\\
\vspace{1em}
$^1$Shanghai Jiao Tong University 
\quad
$^2$AntGroup Research \\
}

\maketitle
\begin{abstract}
Recent studies have combined 3D Gaussian and 3D Morphable Models (3DMM) to construct high-quality 3D head avatars. In this line of research, existing methods either fail to capture the dynamic textures or incur significant overhead in terms of runtime speed or storage space. To this end, we propose a novel method that addresses all the aforementioned demands. 
In specific, we introduce an expressive and compact representation that encodes texture-related attributes of the 3D Gaussians in the tensorial format.
We store appearance of neutral expression in static tri-planes, and represents dynamic texture details for different expressions using lightweight 1D feature lines, which are then decoded into opacity offset relative to the neutral face.
We further propose adaptive truncated opacity penalty and class-balanced sampling to improve generalization across different expressions.
Experiments show this design enables accurate face dynamic details capturing while maintains real-time rendering and significantly reduces storage costs, thus broadening the applicability to more scenarios.

\end{abstract}    
\section{Introduction}
\label{sec:intro}

\begin{figure}
    \centering
    \includegraphics[width=\linewidth]{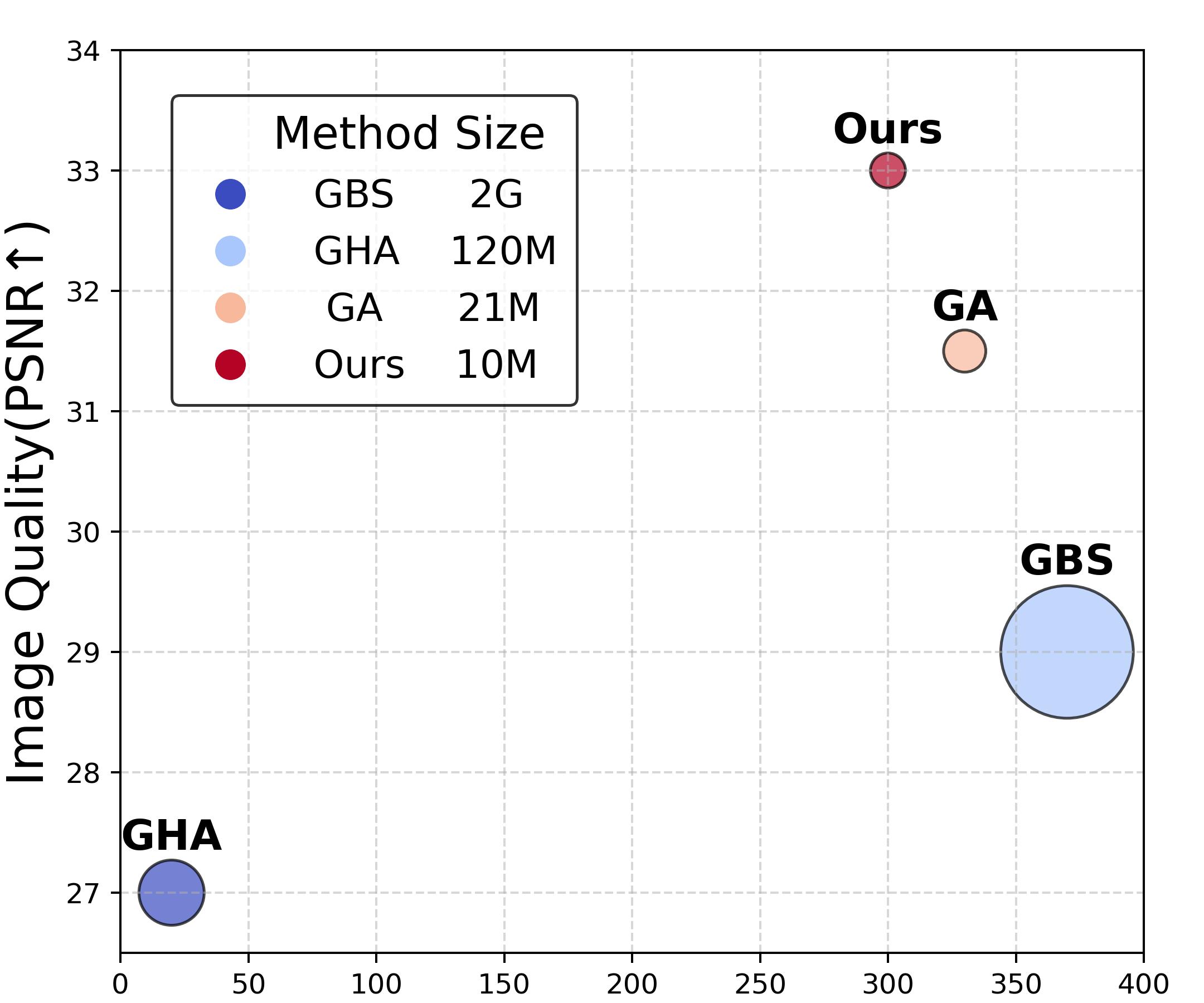}
    \caption{Our method improves rendering quality while ensuring real-time performance and minimum storage. The points radius in the figure is proportional to the square root of the storage.}
    \label{fig:performance}
\end{figure}
Photorealistic 3D head avatars reconstruction is a fundamental research topic in the field of computer graphics and vision, encompassing a variety of applications including films, gaming, and AR/VR etc.
A highly practical 3D head avatar needs to meet various technical requirements: 1) It needs to faithfully render the expressions and movements depicted in the driving signal into high-quality rendered images, including clear texture details, consistent geometric structures, and vivid texture variations under dynamic expressions; 2) It requires extremely high inference and rendering efficiency to meet the demands of tasks that require real-time performance; 3) It needs to meet the requirements of lightweight design to facilitate easy distribution and deployment. 
However, meeting all of the aforementioned requirements remains a significantly challenge.

Early methods are based on 3D face Morphable Models (3DMM) \cite{blanz1999morphable, thies2016face2face} to achieve 3D avatar reconstruction, which can easily control facial expressions using low-dimensional PCA coefficients, but the rendering effect lacks realism.
With the development of NeRF~\cite{mildenhall2021nerf}, some approaches combine NeRF and 3DMM to achieve high-quality, animatable head avatars, but the volumetric rendering process of NeRF requires substantial computation, making it difficult to achieve real-time performance. 
Recently, 3D Gaussian Splatting (3DGS)~\cite{kerbl20233d} has gained widespread attention for its high-quality rendering and real-time performance, with some works attempting to combine 3DGS with 3DMM to achieve fast, realistic, and animatable head avatars.

To enhance realism of head avatar rendering, it is important to model the dynamic facial details that change with expressions, but capturing such details inevitably introduces additional time and memory costs. 
Some methods~\cite{ma20243d} explicitly store a set of Gaussian splats for each blendshape, resulting in significant storage requirements. 
Alternatively, other methods~\cite{xu2024gaussian} employ MLPs to implicitly model dynamic textures, followed by super-resolution techniques to enhance details rendering, but at the cost of not being able to render in real-time.
\revised{Additionally, more parameters and complex networks often lead to overfitting, resulting in poorer generalization ability on novel expressions.}

To address these issues, we propose a novel 3D head avatar modeling method that takes into account both dynamic texture modeling and spatiotemporal efficiency. 
First the efficient 3D Gaussian Splatting rendering pipeline is employed. Then the 3D Gaussian splats are bound to a parametric head mesh, enabling the splats to depict the base motion of the face along with the mesh. Observing that neighboring Gaussian splats share similar appearances and dynamics, hence we store the static texture and dynamic properties in the involved compact tensorial features to reduce spatial redundancy. 
To be specific, the static texture in neutral expression is stored in \textbf{tri-planes} within the canonical space, replacing spherical harmonics of 3DGS. 
For facial dynamics modeling, we store a neural grid representing opacity offsets relative to the neutral face for each blendshape. Then we exploit \textbf{1D feature lines} to depict the dynamic part of facial textures, as experiments show that those features can adequately capture the texture changes caused by facial dynamics, and further reduce the occupied storage. These feature lines are interpolated by blendshape coefficients, and a non-linear MLP decoder outputs the opacity offset.
Benefited from compact architecture, our method enables high-fidelity head avatar with dynamic textures while maintaining storage efficiency and real-time inference.

Compact tensorial representations also reduce the risk of overfitting to training expressions. We further propose two training strategies to improve generalization ability to novel expressions. 
First, as areas which remain consistent to the neutral expression should not have opacity offsets, we propose an adaptive truncated penalty on opacity offset which identifies relatively static mesh triangles in each frame and constrains opacity offset of their corresponding splats to be minimal.
Second, as large-scale expressions are underrepresented in the training set, we propose a class-balanced resampling method: training expressions are clustered, and samples are drawn uniformly from each cluster. 

We conduct experiments on the Nersemble~\cite{kirschstein2023nersemble} dataset, showing that our method accurately reconstructs dynamic facial details and improves rendering metrics. 
Our compact representations require no more than 10MB per subject, making it the most storage-efficient method compared to the state-of-the-art competitors. We achieve 300 FPS, ensuring real-time performance. 
The spatial and temporal efficiency of our approach allows it to be extended to broader application scenarios, such as fast network transmission and real-time rendering in mobile video conferencing.

\section{Related Works}
\subsection{Animatable Head Avatar}
Traditional head avatar methods rely on \textbf{3D} \textbf{M}orphable face \textbf{M}odel~\cite{blanz1999morphable,thies2016face2face} to control face motion by low-dimensional coefficients but lacks rendering realism. Recently, advances in neural rendering has led to combining neural scene representations like NeRF~\cite{mildenhall2021nerf} and 3DGS~\cite{kerbl20233d} with 3DMM to achieve high-quality animatable head avatars~\cite{gafni2021dynamic, qian2024gaussianavatars, gao2022reconstructing, bai2023high, yu2023nofa} by conditioning NeRF or 3DGS on 3DMM coefficients or meshes. 
Further approaches attempt to improve head avatar reconstruction from various perspectives, such as inference speed~\cite{zielonka2023instant, xiang2024flashavatar}, statistical face model~\cite{hong2022headnerf, zhuang2022mofanerf,xu20253d, giebenhain2023learning}, dynamic details~\cite{xu2023avatarmav,chen2023implicit}, sparse-view robustness~\cite{chu2024generalizable, zheng2024headgap} and texture-material disentanglement~\cite{saito2024relightable,xu2024artist}. 
Below, we primarily review 3DGS head avatar methods focused on dynamic facial textures. 
Some approaches~\cite{chen2024monogaussianavatar, xiang2024flashavatar} use MLPs to capture dynamic offsets of gaussian attributes, while others employ CNNs~\cite{gao2022reconstructing, saito2024relightable}. Super-resolution modules are also incorporated to improve the quality of dynamic details rendering~\cite{giebenhain2024npga, xu2024gaussian}. Accurately capturing dynamic details often requires multiple large networks, which highly increase time costs. Methods such as~\cite{dhamo2025headgas, ma20243d} store expression-related dynamic information as additional attributes per Gaussian splat, leading to significant memory usage. By contrast, our approach achieves high-fidelity rendering with dynamic facial details with 10M storage and 300FPS rendering speed.

\begin{figure*}[t!]
    \centering
    \includegraphics[width=\textwidth]{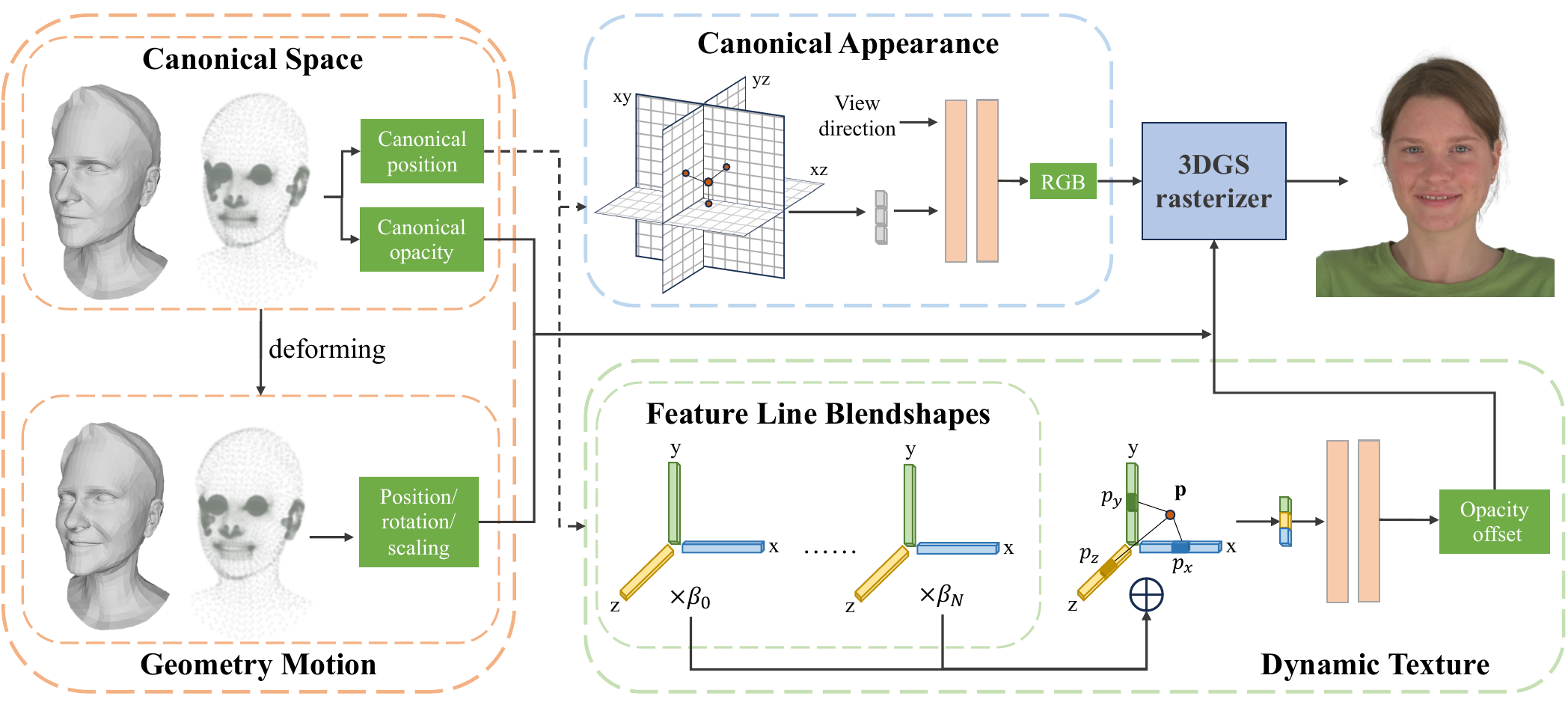}
    \caption{Our goal is to reconstruct 3DGS head avatar with dynamic details, ensuring real-time rendering and minimized storage. We use \yrn{a} \textbf{parametric face mesh} to describe large-scale geometry motions, moving the bound Gaussian splats accordingly. A \textbf{triplane} stores view-dependent appearance in canonical space, while \textbf{1D feature lines} are used for dynamic details per blendshape, allowing interpolation with expression coefficients. Finally, the geometry attributes of the splats, along with the canonical appearance and dynamic details, are combined to render the face image.}
    \label{fig:pipeline}
\end{figure*}

\subsection{Compact 3D Scene Representations for Novel View Synthesis.}

NeRF employs deep MLPs to model scenes, but suffers from slow inference and scalability issues for large, unbounded areas. Some works~\cite{muller2022instant, Chan2022, chen2022tensorf} address this by replacing one large MLP with spatially discrete features and small MLP decoders, improving rendering efficiency. Other methods reduce computation by decomposing large NeRFs into smaller sub-networks~\cite{yang2021learning}.
3DGS is efficient in rendering speed but requires storing a large number of splats for large or complex scene. Recent approaches address this issue using region-based vector quantization~\cite{niedermayr2024compressed}, K-means-based codebooks~\cite{navaneet2023compact3d}, or learned binary masks for each Gaussian.
The memory-intensive spherical harmonics in 3DGS are replaced with more compact neural networks~\cite{lee2024compact, zou2024triplane} for view-dependent radiance. 
Furthermore, some approaches attempt to model dynamic scenes using compact format, such as 2D plane~\cite{fridovich2023k, shao2023tensor4d}, sparse control points~\cite{huang2024sc, jiang2024hifi4g} and parametric curves~\cite{wang2021neural, li2024spacetime}.
In this paper, we leverage tensorial representations(tri-planes and 1D feature lines) to represent static and dynamic appearances.

\section{Method}
Our method takes multi-view face videos as inputs and outputs an animatable head avatar with dynamic textures. 
As shown in Fig.~\ref{fig:pipeline}, our method stores large-scale head motion, head appearance in the canonical space, and dynamic texture variations for each blendshape using three different structures, \textit{i.e.,} mesh, triplane, and feature lines, respectively.
1) Geometry motion bound to mesh: We follow ~\cite{qian2024gaussianavatars} to track FLAME~\cite{li2017learning} mesh for each frame from multi-view images and known camera parameters and calculate splats geometry attributes (position/rotation/scaling) by tracked meshes. 
2) Appearance of neutral face in tri-planes: 
Tri-planes are used to store view-dependent radiance of the canonical 3D face. The features sampled from the tri-planes, and view direction transformed into the canonical space, are fed into a tiny MLP decoder to obtain RGB color.
3) Dynamic details in feature lines:
We utilize a feature line per blendshape to store dynamic textures, which can be interpolated by tracked blendshape coefficients. An MLP decoder then maps the features sampled from the interpolated feature line into opacity offset, which is added to canonical opacity. Finally, the aforementioned gaussian attributes are combined to render the image. 

\subsection{Preliminaries}
\textbf{3DGS.} 3D Gaussian Splatting~\cite{kerbl20233d} enables novel view synthesis of a static scene from multi-view images and camera parameters. A scene is represented by a collection of 3D Gaussian splats, 
and each 3D Gaussian contains the following attributes: position $\mathbf{\mu}\in\mathbb{R}^{3}$, scaling $\mathbf{s}\in\mathbb{R}^{3}$, quaternion $\mathbf{q}\in\mathbb{R}^{4}$, opacity $\mathbf{\alpha}\in\mathbb{R}$ and spherical harmonics $\mathbf{SH} \in \mathbb{R}^{(k+1)^2 \times 3}$ to represent view-dependent color (where $k$ is $\mathbf{SH}$ degree). In our paper, we use $\mathbf{r}\in\mathbb{R}^{3\times3}$ to notate the corresponding rotation matrix to $\mathbf{q}$. Finally the final color for a given pixel is calculated by sorting and blending the overlapped Gaussians.

\noindent\textbf{Gaussian Avatars}~\cite{qian2024gaussianavatars} extends 3DGS from static scene to dynamic avatar by binding Gaussian splats to a tracked head mesh, which is obtained by fitting FLAME (a face shape prior model) parameters to multi-view observations.
Each 3D Gaussian splat is paired with a mesh triangle, with its geometric attributes($\mathbf{\mu'}, \mathbf{r'}, \mathbf{s'}$) defined in the triangle's local coordinate system.
Once optimized, these attributes are fixed, making each splat’s relative position within its triangle fixed, yet allowing global movement as the triangle moves.
For a triangle with three vertices, the average vertex position $T$ is set to origin of local coordinate system. Rotation matrix $R$ is constructed using one edge direction, the triangle normal, and their cross product to represent the orientation transformation from local to global space. 
A scalar $k$ is computed as the mean length of one of the edges and its perpendicular. 
Then the transformation from the local space to the global space is conducted as:
\begin{equation}
    r = Rr', \qquad \mu = kR\mu' + T, \qquad s = ks'.
\end{equation}

\subsection{Appearance in Canonical Space via Triplane}
In 3DGS, 48 out of the total 59 parameters for each Gaussian are used for $\mathbf{SH}$ (3 degrees) to capture view-dependent color. 
Noticing that neighboring splats should have similar appearance, instead of storing 48 $\mathbf{SH}$ parameters per splat, we use a triplane to store implicit encodings, and a tiny MLP decoder to decode the encodings along with view direction to RGB colors, which compresses the model size. 
The triplane \(T\) consists of three orthogonal feature planes aligned with the axes: \(\{T_{xy}, T_{xz}, T_{yz}\} \in \mathbb{R}^{3 \times n_f \times n_f \times n_{d1}}\), where $n_f\times n_f$ is spatial resolution of the 2D feature planes, and $n_{d1}$ is the feature dimension. For any given position \(\mathbf{p}\) in canonical space, the corresponding feature is obtained by projecting \(\mathbf{p}\) onto the axis-aligned planes (the $x-y, x-z$ and $y-z$ planes), interpolating to obtain the features on each feature plane and concatenating the interpolated features, which is formulated as: 
\[
t(\mathbf{p}) = \textit{interp}(T_{xy}, \mathbf{p}_{xy}) \oplus \textit{interp}(T_{xz}, \mathbf{p}_{xz}) \oplus \textit{interp}(T_{yz}, \mathbf{p}_{yz}),
\]
where \(\textit{interp}\) represents bilinear interpolation, \(\oplus\) represents concatenation, and \(\mathbf{p}_{xy}, \mathbf{p}_{xz}, \mathbf{p}_{yz}\) refer to the projected position on each plane.

Note that the triplane is defined in the \textit{canonical space}, which corresponds to the global space of the neutral expression. In contrast, we refer to the global space of the non-neutral expression in each frame as the \textit{deformed space}.
Given a splat with the position $\mathbf{\mu'}$, rotation $\mathbf{r'}$, and scale $\mathbf{s'}$ defined in the local space, it can be transformed into the canonical space and deformed space of the current frame based on the canonical transformation $(R_c, T_c, k_c)$ and the deformed transformation $(R_d, T_d, k_d)$  respectively. 
The local coordinate system can also serve as a bridge to transform between the deformed space and the canonical space. The transformation from view direction in the deformed space (denoted as $\mathbf{v}_d$) to view direction in the canonical space (denoted as $\mathbf{v}_c$) can be formulated as:
\[
\mathbf{v}_c = R_c  R_d^{-1} \mathbf{v}_d.
\]

Finally a tiny MLP decodes $t(\mathbf{p})$ and $\mathbf{v}_c$ into RGB value $\mathbf{c}$, which can be used as the first three components of degree-1 $\mathbf{SH}$ for 3DGS rendering. Additionally, since the majority of facial information is concentrated in the frontal view, with less information available from the side views, we utilize larger feature dimension for $T_{xy}$ and lower dimension for $T_{xz}, T_{yz}$ to achieve storage compression.

\subsection{Dynamic Texture via Feature Line Blendshapes}
Existing researches show that 3D Gaussian texture attributes can be effectively compressed using tri-planes, confirming color consistency in local neighborhoods. We extend this local consistency to dynamic avatars for efficient dynamic details representation. 
For each expression blendshape, we store a separate representation that describes the texture changes (opacity offsets) relative to the neutral expression induced by that expression. 
Since the FLAME model has 100 PCA blendshapes, storing 2D planes or 3D tensors for each blendshape incurs excessive memory consumption. 
To enable a compact representation for dynamic textures, we use lightweight 1D feature lines to encode the texture changes for blendshape $i$, denoted as $(L_{x}^{i}, L_{y}^{i}, L_{z}^{i}) \in \mathbb{R}^{3 \times n_{d2} \times n_s}$, where $n_{s}$ represents the length of the 1D feature line, and $n_{d2}$ represents the feature dimension. 
We interpolate feature lines by tracked blendshape coefficients $\beta_j\in \mathbb{R}^{n_b}$ to obtain specific feature line for frame $j$, where $n_b$ is blendshape count, which can be formulated as: 
\[
l^{j}_b = \sum_{i=0}^{n_b} \beta_j^i * (L_x^i, L_y^i, L_z^i).
\]

In addition to the linear blendshapes, FLAME incorporates a nonlinear quaternion jaw rotation to describe large-scale jaw movements. To unify the linear basis with the nonlinear rotation, we follow the method proposed in \cite{li2023posevocab}, extracting linear jaw rotation bases \(\{\mathbf{q}_{k} : k \in \{0, \ldots, n_j\}\}\) from jaw rotations in training videos via farthest point sampling. And we store a feature line $(L_x^k, L_y^k, L_z^k)$ for each jaw basis. 
We follow \cite{huynh2009metrics} to calculate the distance between jaw rotation of frame $j$ and the $k$-th jaw rotation basis which can be formulated as $d(j, k) = 1 - |\mathbf{q}_j^T \mathbf{q}_k|$($\mathbf{q}_j$ and $\mathbf{q_k}$ are unit quaternion). And the jaw feature lines are interpolated using inverse distance weighting to calculate the feature line of frame $j$, formulated as:
\[
l^j_r = \sum_{k=0}^{n_j}\beta_j^{k} * (L_x^{k}, L_y^{k}, L_z^{k}), \qquad \beta^{k}_j = \frac{1-d(j, k)}{\sum_{k=0}^{n_j}(1 -d(j,k))}.
\]

Similar to the triplane, the opacity offset features of a given position $\mathbf{p}$ in frame $j$ is calculated by projecting $\mathbf{p}$ onto the $x, y$ and $z$ axes, interpolating and concatenating the interpolated features, formulated as:
\[
l^{j}(\mathbf{p}) = \textit{interp}(l^j_x, \mathbf{p}_x) \oplus \textit{interp}(l^j_y, \mathbf{p}_y) \oplus \textit{interp}(l^j_z, \mathbf{p}_z).
\]
This projection and interpolation process is applied to both the expression blendshape feature line $l^{j}_b$ and the jaw rotation feature line $l^{j}_r$.

We finally utilize a tiny MLP $\theta$ to decode interpolated $l_b^j(\mathbf{p})$ and $l_r^j(\mathbf{p})$ into opacity offset $\Delta\alpha$, which will be added to the canonical opacity $\alpha_c$ (opacity of the neutral experssion). The final opacity is calculated as:
\[
\alpha = \alpha_{c} + \Delta\alpha = \alpha_{c} + \theta(l_b^j(\mathbf{p}), l_r^j(\mathbf{p})).
\]

The facial motion caused by facial expressions are primarily concentrated in the leading components of FLAME PCA blendshapes.  Using only the leading expression coefficients achieves similar results, allowing us to reduce the number of feature lines, thereby reducing storage and accelerating calculation. 

\subsection{Training}
\noindent\textbf{Class-balanced Sampling.} 
Most images in the training set show minor facial movements, with only few displaying large expressions, making it challenging to accurately reconstruct these. Since the data distribution is unknown, simply oversampling large expressions may bias against smaller ones. We propose a resampling method to ensure that various expressions are sampled evenly during training.

First, we measure the similarity between two frames by comparing the differences in vertex displacements of the FLAME mesh. 
Given that different regions of the face exhibit varying degrees of motion, \textit{e.g.,} the lips moving significantly more than the eyes, we empirically increase weights of the eyes vertices. 
We denote the FLAME mesh for frame $i$ as $M_i$, and the similarity score between frame $i$ and frame $j$ is calculated as $dist(i, j) = ||M_i - M_j||_2^2 * \mathbf{w}$, where $\mathbf{w}$ denotes the weight of each vertex. We set $\mathbf{w}=2$ for eye vertices and $\mathbf{w}=1$ for other vertices.
Next, we perform spectral clustering on the similarity matrix to categorize all frames into $n=16$ classes. 
Finally, we utilize a uniform sampling probability distribution between categories and equal probability within each category.

\noindent\textbf{Adaptive Truncated Opacity Offset Penalty.}
\yrn{The} feature lines store the opacity offset of each blendshape relative to the neutral face, hence \yrn{the} offsets in relatively static regions \yrn{are} ideally zero. Therefore, we calculate triangle translations $\overline{\mathbf{t}}$\yrn{, \emph{i.e.}, the displacement between the deformed mesh and the neutral face mesh,} 
and empirically set a threshold $\tau$ to identify whether the triangle is ``static" or ``dynamic". Then opacity offsets of splats bound to ``static" triangles are constrained to be \yrn{close to $0$}.
This penalty helps decouple static textures from dynamic details, enhancing generalization to unseen expressions, \yrn{formulated as}:
\begin{equation}
        \mathcal{L}_{op} = \lambda_{op}|\Delta\alpha| * w_{op}, \quad w_{op} = 
        \begin{cases}
        \yrn{0},\quad \yrn{\text{if }} |\overline{\mathbf{t}}| > \tau, \\
        \yrn{1},\quad \yrn{\text{if }} |\overline{\mathbf{t}}| <= \tau. \\
        \end{cases}
\end{equation}

\noindent\textbf{Loss Function.} 
We use the L1 loss and D-SSIM loss between the rendered images and the ground truth images as image supervision, which can be formualted as:
\[
\mathcal{L}_{image} = (1-\lambda)\mathcal{L}_1 + \lambda\mathcal{L}_{\text{D-SSIM}}.
\] 
Assuming that the Gaussian splats should roughly conform to the mesh and be similar in size with the bound triangles, we follow methods proposed by \cite{qian2024gaussianavatars} to employ a position loss and a scale loss to prevent splats from being excessively far from the mesh or excessively large: 
\begin{equation}
        \mathcal{L}_{geom} =\lambda_{pos}\mathcal{L}_{pos} + \lambda_{scale}\mathcal{L}_{scale}.
\end{equation}
The \yrn{total} training loss can be formulated as:
\[
\mathcal{L} = \mathcal{L}_{image} + \mathcal{L}_{geom} + \mathcal{L}_{op},
\]
where $\lambda=0.2$, $\lambda_{pos}=0.01$, $\lambda_{scale}=1$, \yrn{and} $\lambda_{op}=1$. 
\definecolor{best}{rgb}{0.894,0.937,0.862}
\definecolor{second}{rgb}{0.992,0.952,0.815}

\begin{figure*}[ht!]
    \centering
    \includegraphics[width=0.93\textwidth]{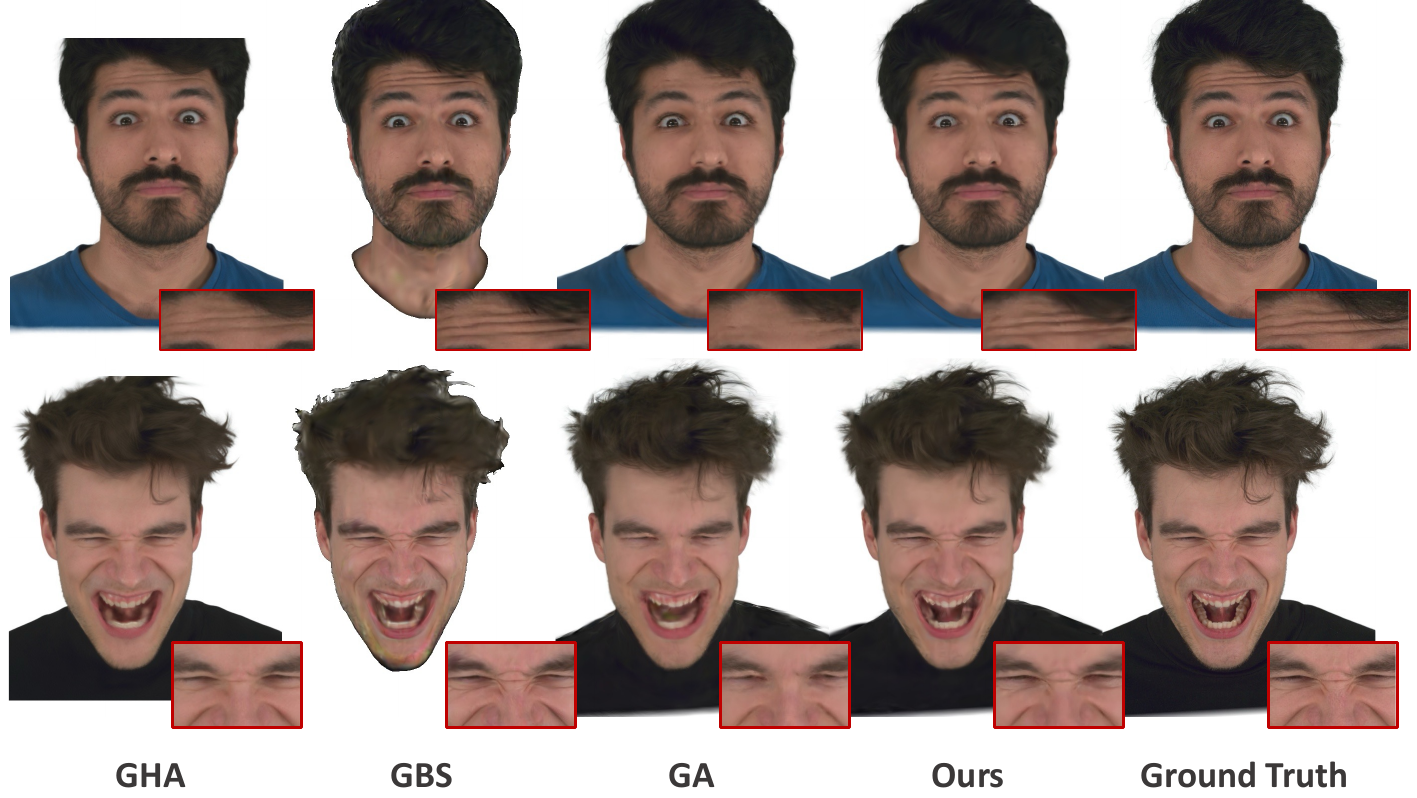}
    \caption{Qualitative comparison with baseline methods on novel view synthesis task. }
    \label{fig:nvs}
\end{figure*}
\begin{table*}[t!]
    \centering
    \resizebox{0.9\textwidth}{!}{
        \begin{tabular}{l|ccc|ccc|cc} 
        \hline
        & \multicolumn{3}{c|}{Novel View Synthesis} & \multicolumn{3}{c|}{Self-Reenactment} & \multicolumn{2}{c}{Performance} \\ \hline
        Method & PSNR$\uparrow$ & SSIM$\uparrow$ & LPIPS$\downarrow$ & PSNR$\uparrow$ & SSIM$\uparrow$ & LPIPS$\downarrow$ & Storage & FPS  \\ \hline
        GA     &\colorbox{second}{31.4702}&0.9489&\colorbox{second}{0.05144}&\colorbox{second}{27.2678}&0.9230&\colorbox{best}{0.06668}&\colorbox{second}{21M}      & \colorbox{second}{330}              \\ 
        GHA    & 26.9932  & 0.9347 &\colorbox{best}{0.04905}  & 22.7397  &0.8895   & 0.07995    &120M  & 20                \\ 
        GBS&  28.8966  &  \colorbox{second}{0.9500}  & 0.06311  & 25.9797 & \colorbox{best}{0.9270} &0.08099    &2G     &    \colorbox{best}{370}           \\ 
        \textbf{Ours}  &  \colorbox{best}{32.9688}&  \colorbox{best}{0.9506} & 0.05940  & \colorbox{best}{28.0688} & \colorbox{second}{0.9259}  &  \colorbox{second}{0.07724}    &\colorbox{best}{10M}  & 300                   \\ \hline
    \end{tabular}
    }

    \caption{Quantitative comparison with baselines. \colorbox{best}{GREEN} indicates the best of all methods. \colorbox{second}{YELLOW} indicates the second.}
    \label{tab:comparison}
\end{table*}
\section{Experiments}

\subsection{Settings and Dataset}
We conduct experiments on nine individuals from the nersemble dataset, collecting a total of 11 video segments of each subject from 16 different viewpoints. Each participant performs 10 distinct expressions and emotions as instructed, followed by a free performance in the last video segment. The videos are downsampled to a resolution of 802 × 550. We utilize the FLAME coefficients and camera parameters provided in GA~\cite{qian2024gaussianavatars}, including shape $\beta$, translation $\mathbf{t}$, pose $\mathbf{\theta}$, expression $\psi$, and vertex offset $\Delta \mathbf{v}$ in the canonical space.

We compare the experimental results across three tasks: 1) Novel View Synthesis: 15 out of 16 viewpoints are used for training, while the remaining viewpoint is reserved for testing; 2) Self-Reenactment: Testing is conducted using videos of the same individual showcasing unseen poses and expressions from all 16 viewpoints; 3) Cross-Identity Reenactment: An avatar is driven by the motions and expressions of other individuals. We use free performance sequences for testing of task 2 and task 3.
\subsection{Implementation Details}
We implemented \yrn{our approach} using PyTorch and trained for 600,000 iterations using the Adam optimizer for each subject.
Both the triplane and feature lines consist of two components: \yrn{a} neural grid feature and \yrn{a} MLP decoder. We train triplane and feature line blendshapes with the same learning rates, setting  learning rate of \yrn{the} feature to $2e-3$ and the MLP learning rate to $1e-4$. 
Each plane of the triplane is $128\times128$ in size, with the feature dimension of the $T_{xy}$ being $32$, and the feature dimensions of the $T_{xz}, T_{yz}$ planes set to $16$. The MLP decoder following triplane consists two-layer with 128 dimensions per layer, using ReLU as the activation function. We apply position encoding to improve the resolution of view direction.
We assign feature lines with a spatial resolution of 64 and feature dimension of 32 to the first 80 PCA expression bases and 16 key jaw rotation bases. The decoder for the feature lines is a two-layer MLP with 128 dimensions per layer. 
The triplane requires 4.05M of storage, the feature line requires 2.41M, and the other Gaussian attributes (including position/rotation/scaling and canonical opacity) average 3.25M per subject.

\subsection{Comparison}
\begin{figure*}
    \centering
    \includegraphics[width=0.93\textwidth]{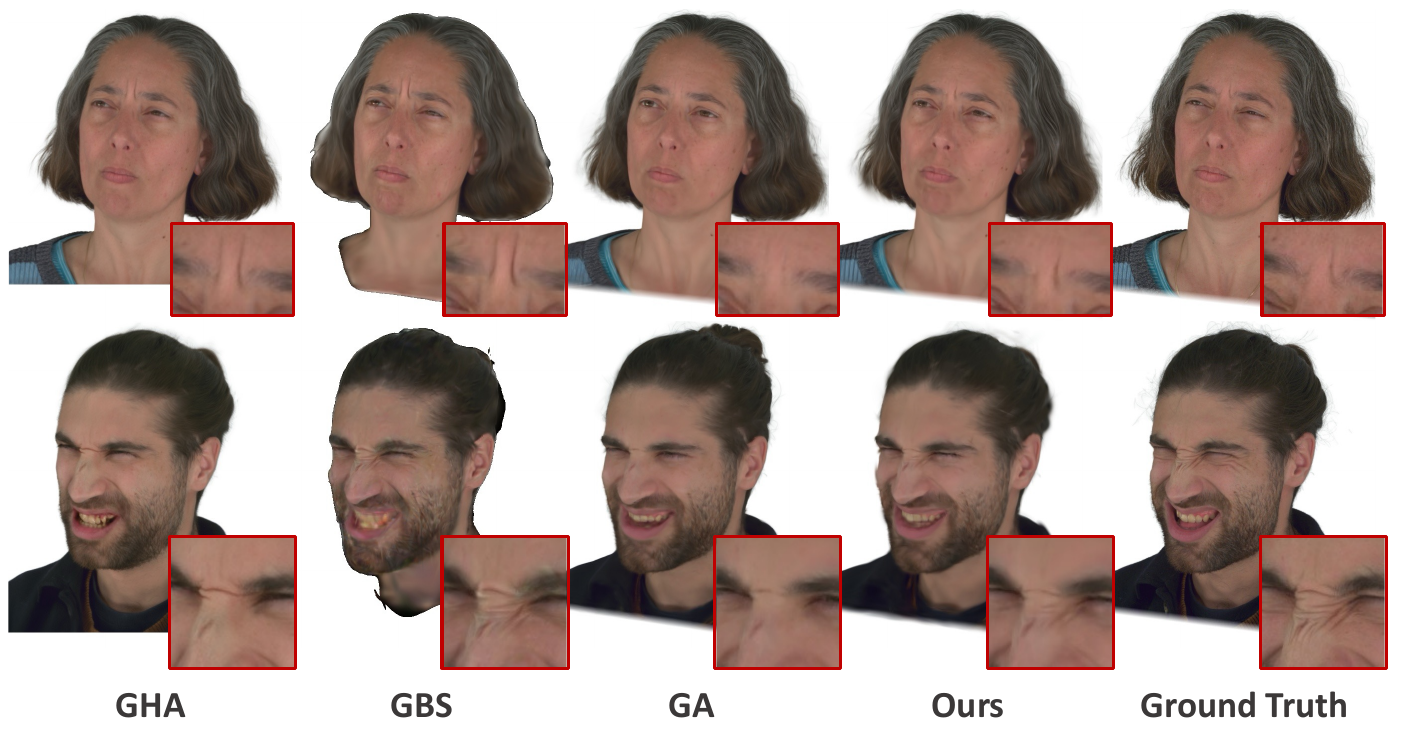}
    \caption{Qualitative comparison with baseline methods on self-reenactment task.}
    \label{fig:self-reenact}
\end{figure*}
\begin{table*}[h]
    \centering
    \resizebox{0.9\textwidth}{!}{
    \begin{tabular}{cccc|ccc|ccc}
    \hline
    \multicolumn{4}{c|}{Components} & 
    \multicolumn{3}{c|}{Novel View Synthesis} & 
    \multicolumn{3}{c}{Self-Reenactment}\\
    \hline
        trip& fl& penalty& resample & PSNR$\uparrow$ & SSIM$\uparrow$ & LPIPS$\downarrow$ & PSNR$\uparrow$ & SSIM$\uparrow$ & LPIPS$\downarrow$ \\ 
    \hline
         &&&\checkmark & 33.4206& 0.9684& 0.02968& 31.2487& 0.9561& \colorbox{second}{0.03578}\\
         \checkmark &&&\checkmark& 33.4337& 0.9687& 0.03090&31.3246& \colorbox{second}{0.9585}& 0.03792 \\
         &\checkmark&\checkmark&\checkmark& \colorbox{second}{35.3876}& \colorbox{best}{0.9739}& \colorbox{best}{0.02497}&\colorbox{second}{31.4366}& 0.9577& \colorbox{best}{0.03564} \\
          \checkmark&\checkmark&\checkmark&&35.2652& \colorbox{second}{0.9737}&\colorbox{second}{0.02590}&31.4288 & 0.9580&0.03585 \\   \checkmark&\checkmark&&\checkmark&\colorbox{best}{35.5509}&0.9734&0.02597&31.4042&0.9580&0.03790\\
          \checkmark&\checkmark&\checkmark&\checkmark& 35.1644& 0.9715& 0.02605&\colorbox{best}{31.6446}& \colorbox{best}{0.9591}& 0.03580
\\
    \hline
    \end{tabular}
    }
    \caption{Ablation Study on subject \#306. ``trip" and "fl" refer to neutral texture triplane and feature line blendshapes respectively. ``penalty" indicates opacity offset penalty. ``resample" indicates class-balanced sampling. 
    \colorbox{best}{GREEN} indicates the best and \colorbox{second}{YELLOW} indicates the second.}
    \vspace{-0.1in}
    \label{tab:ablation}
\end{table*}

We compare three baselines with our method.

\noindent\textbf{1) GA}~\cite{qian2024gaussianavatars} defines Gaussian splats in the relative coordinate system of the mesh triangles, allowing splats to move with the mesh. However, GA uses the same set of Gaussian splats for all expressions, limiting its ability to capture expression-specific details.

\noindent\textbf{2) GBS}~\cite{ma20243d} optimizes a set of 3D Gaussians splats per blendshape, which can be interpolated linearly using 3DMM coefficients, but at the cost of significant storage. Moreover, extensive parameters may lead to overfitting to training expressions and simple linear interpolation of splats attributes may introduces artifacts in large expressions.

\noindent\textbf{3) GHA}~\cite{xu2024gaussian} uses coarse guide meshes and two MLPs to predict geometry and color offsets, and converts 3D Gaussian features into RGB images via super-resolution. GHA achieve highly detailed rendering, but fail to achieve real-time rendering speed.

Different from GA, GHA and our method, GBS inputs monocular videos and reconstructs the head but excluding the clothing and shoulders. To ensure a fair comparison, we use multi-view videos and the FLAME parameters tracked following GA to serve as input of GBS, and segment the clothing area using face parsing, excluding it from the quantitative metrics. We employ PSNR, SSIM, and LPIPS~\cite{zhang2018unreasonable} as quantitative metrics while listing the required storage and FPS, which can be found in Table~\ref{tab:comparison}. The storage size does not include pre-tracked 3DMM parameters. The FPS is tested on NVIDIA RTX4090 GPU. Our method occupies the minimum storage, ensuring real-time rendering speed, while also improving PSNR metrics on the tasks of novel view synthesis and self-reenactment.

Fig.~\ref{fig:nvs} demonstrates the effectiveness of our method on novel view synthesis task. Compared to GA and GHA, our approach better reconstructs dynamic textures generated by significant expressions, such as the forehead wrinkles in the first row and the wrinkles around the right eye in the second row. 
GBS linearly interpolates all Gaussian attributes using blendshape coefficients, which can result in artifacts such as the left chin in the second row where exists large-scale non-linear motion and neck in the first row. In contrast, our method provides more stable modeling while maintaining comparable dynamic details. Noted that GBS requires 2GB storage, whereas ours requires only 10MB.

\begin{figure*}
    \centering
    \includegraphics[width=0.93\textwidth]{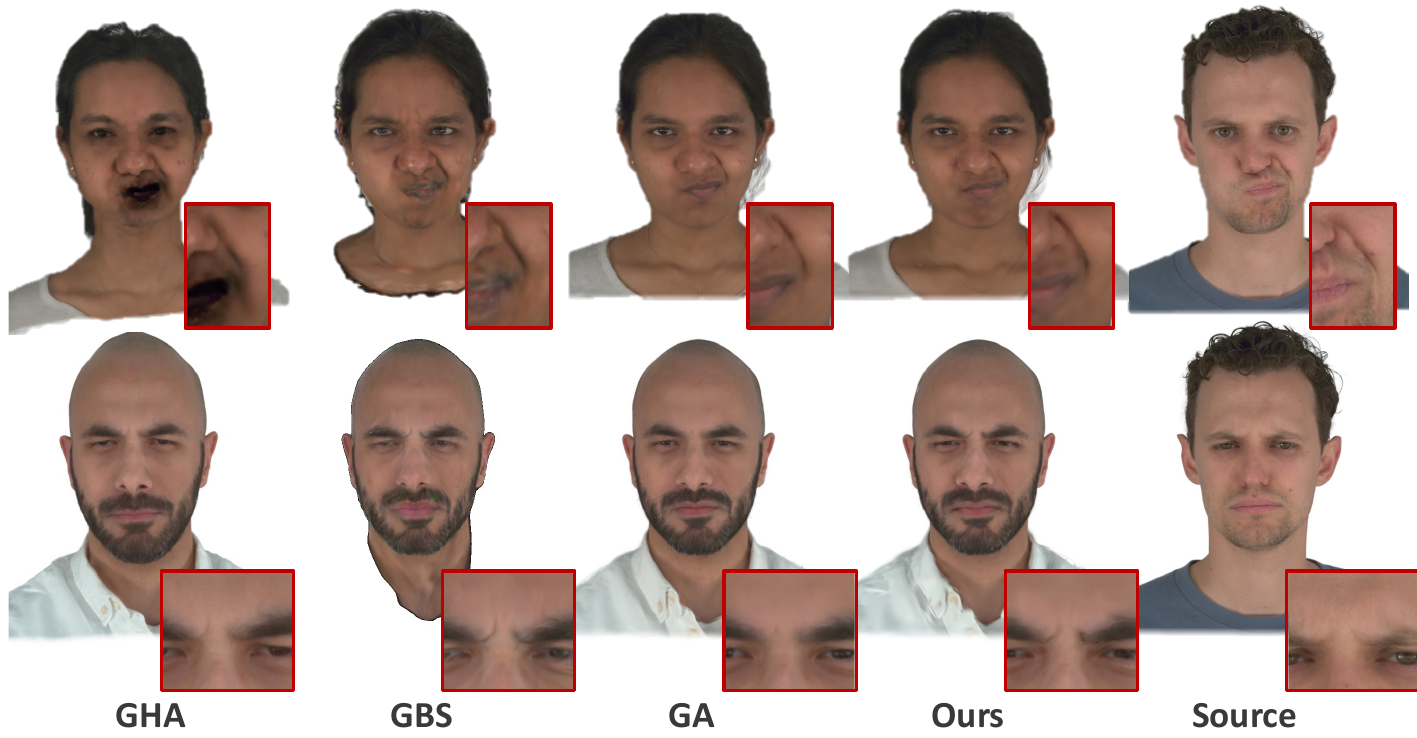}
    \caption{Cross-identity reenactment of head avatars. We use the expression and pose of the source subject on the far right to drive the character on the left.}
    \vspace{-0.1in}
    \label{fig:cross}
\end{figure*}

Fig.~\ref{fig:self-reenact} shows the qualitative comparison results for the self-reenactment task. Our method enables avatar rendering with dynamic textures and less artifacts
while avoiding overfitting of the wrinkles, such as the frown lines in the first row of Fig.~\ref{fig:self-reenact}.
Fig.~\ref{fig:cross} illustrates the performance of our method in the cross-reenactment task, showcasing the generation of distinct wrinkle effects and identity preservation.

\subsection{Ablation Study}
We conduct ablations on subject \#306 to evaluate the proposed components. Noted that novel view synthesis task involves novel view on training expressions, while self-reenactment involves novel expressions on $15$ training views and one novel view.

\noindent\textbf{Tensorial representations.} In Table~\ref{tab:ablation}, the first two row\yrn{s} show compact triplanes help avoid overfit\yrn{ting}, improving metrics on unseen expressions, while it may blur high-frequency details, leading to a \yrn{worse} LPIPS score. 
The first and third rows show that involving feature lines to model dynamic textures enhance\yrn{s} both novel view and novel expressions rendering.

\begin{figure}[t!]
    \centering
    \includegraphics[width=0.8\linewidth]{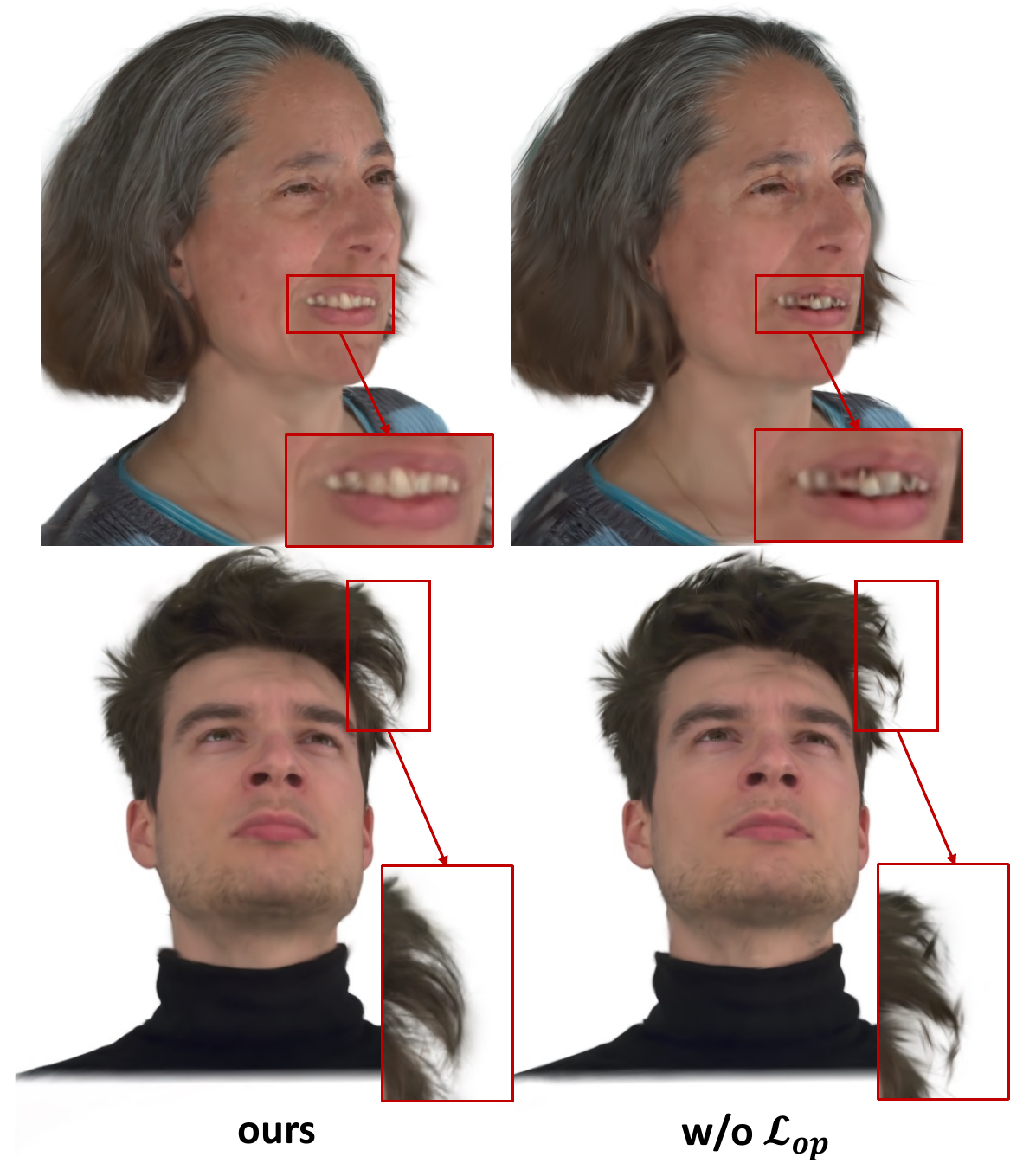}
    \vspace{-0.15in}
    \caption{$L_{op}$ helps prevent artifacts in self-reenactment. }
    \vspace{-0.15in}
    \label{fig:norm}
\end{figure}

\noindent\textbf{Class-balanced Sampling.} We conduct ablations on the class-balanced sampling, as shown in the \yr{4th and 6th} rows of Table~\ref{tab:ablation}. Frames with large expressions are less frequent in the training data compared to frames with small motion. The sampling prevents the model from overfitting to relatively static frames, thereby improving generalization to unknown expressions distribution, \yrn{achieving better self-reenactment performance}.

\noindent\textbf{Adaptive truncated opacity offset penalty.}
Ablation studies are conducted to evaluate the adaptive truncated opacity offset penalty, as shown in \yrn{5th and 6th} rows of  Table~\ref{tab:ablation}. This penalty helps disentangle canonical appearance with dynamic details, thus improving robustness on unseen expressions. \yrn{Moreover,} without this \yrn{penalty}, floaters may appear around the hair, and artifacts may occur in the teeth area during expression changes, as shown in Fig.~\ref{fig:norm}.

\subsection{Conclusion and Limitations}
In this paper, we propose a 3DGS head avatar modeling method that balances dynamic details capturing with real-time performance and low storage. Compact tensorial features (2D triplane for canonical appearance, and lightweight 1D feature lines for dynamic details) allow for accurate appearance modeling. Adaptive opacity offset penalty and class-balanced training helps prevent from overfitting to the training expressions.
Experiments demonstrate that our approach not only enhances the rendering quality but also maintains real-time rendering speed and minimal storage, making it suitable for a wide range of practical applications.

Our method has some limitations, such as reliance on the tracked mesh thus unable to handle complex hairstyles or topological changes in the mouth interior. Our method do not decouple material and lighting, so avatar relighting is not feasible, which we intend to explore in the future.

\section*{Acknowledgements}
This work was supported by National Natural Science Foundation of China (72192821, 62302297 and 62472282), Shanghai Sailing Program(22YF1420300), Young Elite Scientists Sponsorship Program by CAST (2022QNRC001), YuCaiKe [2023](231111310300) and the Fundamental Research Funds for the Central Universities(YG2023QNA35). This work was also supported by AntGroup Research Intern Program.






{
    \small
    \bibliographystyle{ieeenat_fullname}
    \bibliography{main}
}

\clearpage
\setcounter{page}{1}
\maketitlesupplementary

\section{Implementation Details}
\begin{figure}[ht]
    \centering
    \includegraphics[width=\linewidth]{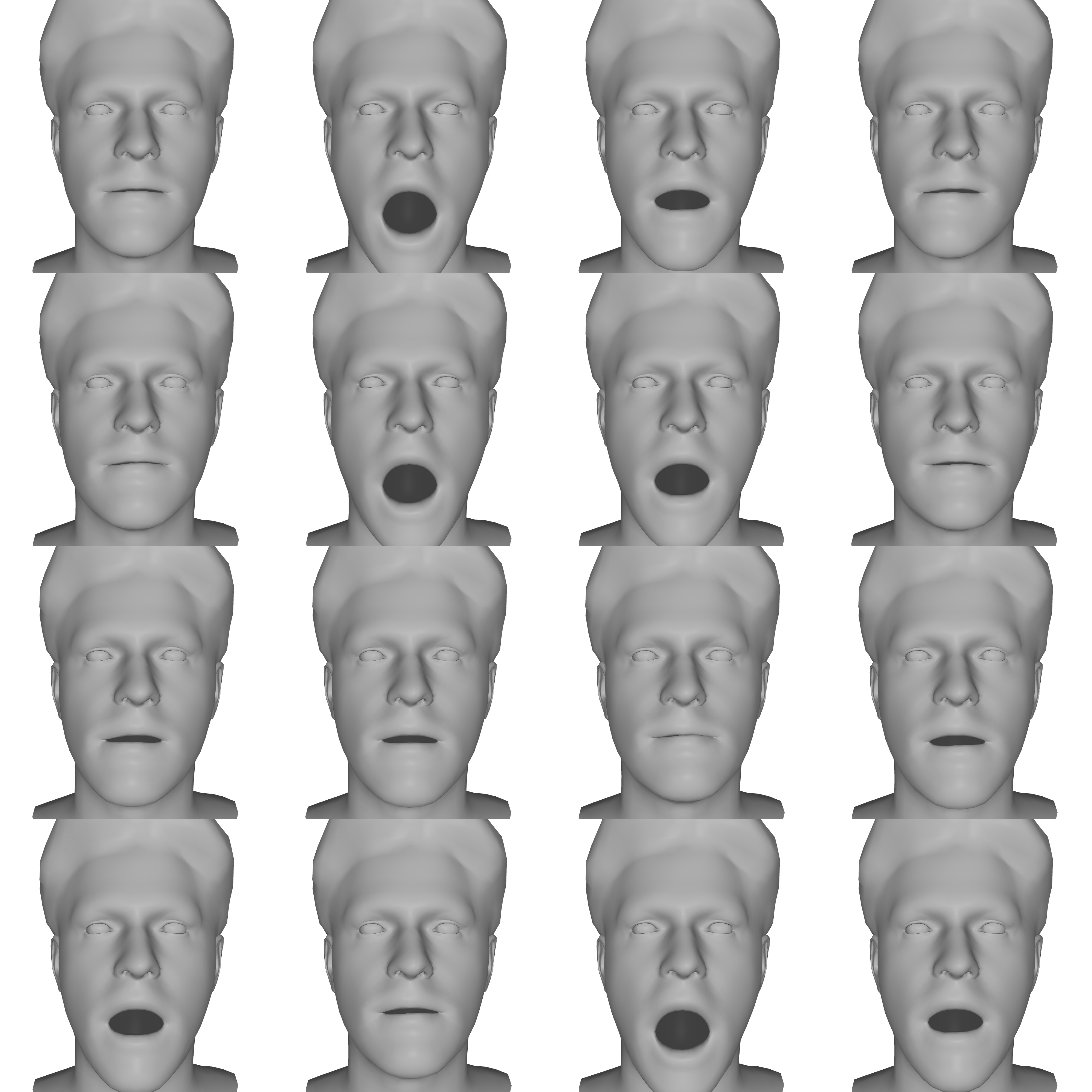}
    \caption{Basis jaw rotation extracted from all frames from videos via farthest point sampling of subject\#074.}
    \label{fig:jaw_pose}
\end{figure}

\noindent\textbf{Jaw Pose Linear Bases.}
We use farthest point sampling to extract linear jaw pose bases from jaw poses of each frames in videos, in order to unify dynamic textures due to linear blendshape and non-linear jaw rotation to linearly-interpolated feature lines.
The jaw pose linear bases of id \#074 are shown in Fig~\ref{fig:jaw_pose}. 

\begin{figure}[ht]
    \centering
    \includegraphics[width=\linewidth]{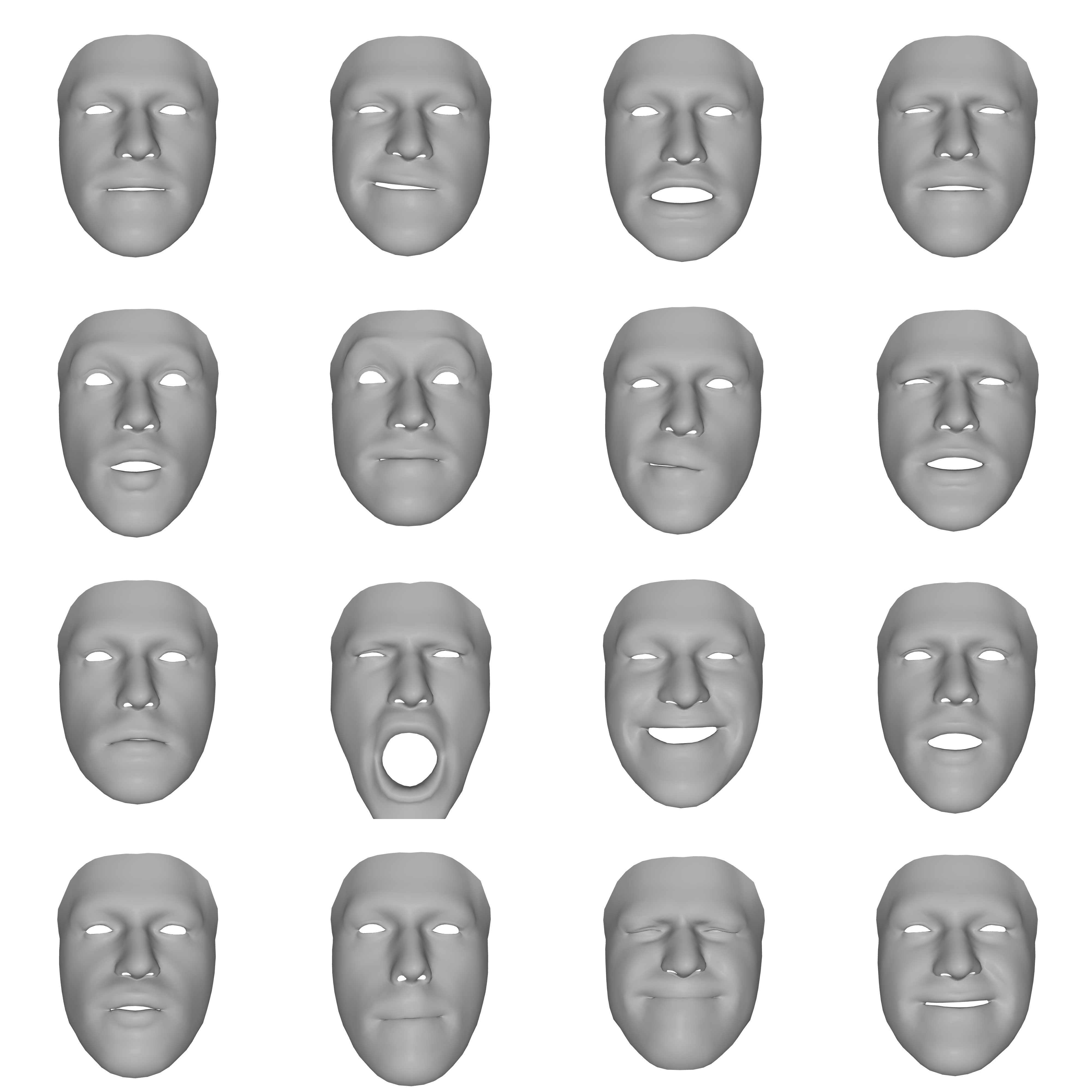}
    \caption{Cluster center for expression balanced sampling of subject\#074.}
    \label{fig:bs-class}
\end{figure}

\noindent\textbf{Class-balanced Sampling.} 
The training frames are clustered into $16$ categories and we evenly sample from $16$ categories to ensure no bias toward expressions with less motion.
We show the cluster center of subject \#074 in Fig~\ref{fig:bs-class}.

\noindent\textbf{Acceleration.} In our experiments, FLAME meshes are generated during the initialization stage to reduce time consumption during inference. Since dynamic textures are primarily concentrated on the face, the spatial bounds of feature lines are set around the face. Splats outside these bounds are excluded when calculating the opacity offset to further accelerate inference.

\section{Comparison Details}
\subsection{Baselines}
We conduct comparative experiments with three baseline methods: GA, GHA, and GBS. To ensure fair comparisons, we align the inputs, including image resolution and pre-tracked mesh. 

\noindent\textbf{GHA.} Both GHA and our method utilize multi-view videos from the Nersemble dataset, but the input resolutions differ. GHA processes 2K resolution images, while our input images are downsampled by a factor of four. To ensure fairness in testing, we first downsampled the 2K images by four times and then upsampled them back to their original size as the input for GHA.
The FPS of GHA is tested by rendering 1024*1024 resolution images.

\noindent\textbf{GBS.} GBS is a monocular facial video reconstruction method, requiring monocular metrical tracker~\cite{zielonka2022towards} to regress FLAME(2020 version, with two additional expression bases for describing closed eyes) coefficients and camera parameters from the images, which serve as the model's input. In our approach, the input consists of multi-view videos along with camera parameters and tracked FLAME(2023 version) coefficients. 

\begin{figure*}
    \centering
    \includegraphics[width=\textwidth]{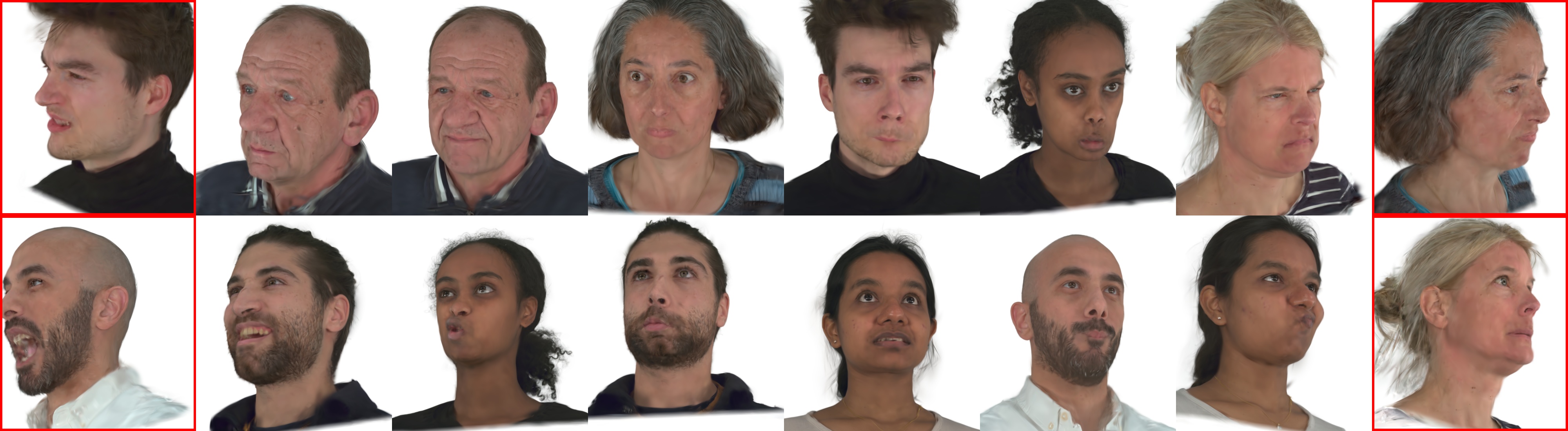}
    \caption{Rendering results of extreme viewpoints and expressions. Extrapolated viewpoints are in the red box.}
    \label{fig:extreme}
\end{figure*}

To ensure fairness of comparisons, we concatenate the multi-view videos into a single video and fit FLAME2020 coefficients to approximate FLAME2023 multi-view tracked meshes instead of monocular metrical tracker, serving as inputs of GBS.
We optimize the FLAME 2020 parameters by calculating the mesh vertex positions loss. Note that the parameters output by the metrical tracker do not include the hair offset or neck motion, so we calculate the loss using only the facial region vertices. First, we compute the shape coefficients using the neutral expression, then regress the expression coefficients, eye rotation and jaw rotation for each frame in an iterative manner.

\subsection{Dataset}
We test our method on nine subjects (074, 104, 165, 175, 210, 218, 264, 302, 304) from Nersemble datasets. The free performace sequences are used to evaluate the effects of self-reenactment, which may contain some frames where the tongue is sticking out. As our method and compared baselines do not focus on mouth interior modeling, we exclude these frames from evaluation.

\section{More Experiments}

\noindent\textbf{NeRF head avatar.} INSTA~\cite{zielonka2023instant} is a NeRF based head avatar method which enables fast training and inference. INSTA relies on FLAME mesh to guide NeRF to move correctly, which warps points according to the nearest mesh triangle directly. 
\begin{table}[h]
    \centering
    \begingroup
    \setlength{\tabcolsep}{1pt}
    \resizebox{\linewidth}{!}{
        \begin{tabular}{l|ccc|ccc|cc} 
        \hline
        & \multicolumn{3}{c|}{Novel View Synthesis} & \multicolumn{3}{c|}{Self-Reenactment} & \multicolumn{2}{c}{Performance} \\ \hline
        Method & PSNR$\uparrow$ & SSIM$\uparrow$ & LPIPS$\downarrow$ & PSNR$\uparrow$ & SSIM$\uparrow$ & LPIPS$\downarrow$ & Storage & FPS  \\ \hline
        INSTA     &27.97&0.92&0.11&27.50&0.92&0.103&53M      & 25             \\ 
        \textbf{Ours}  &  32.97&  0.95 & 0.059  & 28.07 & 0.93  &  0.077    &10M  & 300                   \\ \hline
    \end{tabular}
    }
    \endgroup
    \caption{Quantitative comparison with INSTA. }
    \label{tab:insta}
\end{table}

\noindent\textbf{Offset on Opacity or Other Attributes.}
We tested adding offsets on position/rotation/scale of Gaussians to model face dynamics, but found this design leads to inferior performance of novel expression synthesis (Tab.~\ref{tab:geo_offset} right), since it increases the model’s degrees of freedom, making it prone to overfitting to training expressions, with weaker generalization to novel expressions.

\begin{table}[h]
    \centering
    \resizebox{\linewidth}{!}{
    \begin{tabular}{l|ccc|ccc}
         \hline
        & \multicolumn{3}{c|}{Novel View Synthesis} & \multicolumn{3}{c}{Self-Reenactment} \\ \hline
        - & PSNR$\uparrow$ & SSIM$\uparrow$ & LPIPS$\downarrow$ & PSNR$\uparrow$ & SSIM$\uparrow$ & LPIPS$\downarrow$ \\ \hline
        Opacity & 35.16& 0.97 & 0.026 & 31.64 & 0.96 & 0.036\\
        Others & 36.39& 0.97 & 0.018 & 30.97 & 0.96 & 0.030\\ 
        \hline
    \end{tabular}
    }
    \vspace{-8pt}
    \caption{\yr{Ablation study of} opacity offsets on subject \#306. ``Others`` refers to position/rotation/scale offsets. }
    \label{tab:geo_offset}
\end{table}

\noindent\textbf{Extreme Viewports and Expressions.}
We conduct experiments to validate robustness of our method on extreme viewports and expressions.
We interpolate $16$ new turntable viewpoints from $16$ training views, randomly select from $8$ subjects and generate expression coefficients by sampling the first $8$ dimensions of PCA space, which can be found in Figure~\ref{fig:extreme}.

\section{Ethical Considerations}
The generation of artificial portrait videos using our method poses risks, including the spread of false information, and erosion of trust in media credibility. These issues could have profound societal implications. Addressing this challenge requires developing reliable techniques to identify and verify authentic content.


\end{document}